\documentclass[runningheads]{llncs}
\usepackage{wrapfig}
\usepackage{floatflt}
\usepackage{comment}
\usepackage{amsmath,amssymb} 
\usepackage{mathtools} 
\usepackage{multirow}
\newtheorem{manualtheoreminner}{Principle}
\newenvironment{manualtheorem}[1]{%
  \renewcommand\themanualtheoreminner{#1}%
  \manualtheoreminner
}{\endmanualtheoreminner}

\usepackage{makecell}
\usepackage[ruled,vlined]{algorithm2e}
\newcounter{algoline}

\usepackage{algpseudocode}
\usepackage[accsupp]{axessibility}  
\usepackage[pagebackref,breaklinks,colorlinks]{hyperref}
\newcommand{\ie}{\textit{i}.\textit{e}.}
\newcommand{\eg}{\textit{e}.\textit{g}.}
\usepackage[capitalize]{cleveref}
\usepackage[final]{pdfpages}
\crefname{section}{Sec.}{Secs.}
\Crefname{section}{Section}{Sections}
\Crefname{table}{Table}{Tables}
\crefname{table}{Tab.}{Tabs.}

\begin{document}
\pagestyle{headings}
\mainmatter
\def\ECCVSubNumber{4888}  

\title{Class Is Invariant to Context and Vice Versa: On Learning Invariance for Out-Of-Distribution Generalization} 

%
\author{Jiaxin Qi\inst{1} \and
Kaihua Tang\inst{1}\protect\thanks{Corresponding author.} \and
Qianru Sun\inst{2} \and
Xian-Sheng Hua\inst{3} \and
Hanwang Zhang\inst{1}}

%
\institute{Nanyang Technological University \and
Singapore Management University \and Damo Academy, Alibaba Group \\
{\tt\small jiaxin003@e.ntu.edu.sg,
kaihua.tang@ntu.edu.sg,
qianrusun@smu.edu.sg, xshua@outlook.com,
hanwangzhang@ntu.edu.sg}}
\titlerunning{Learning Invariance for OOD} 
\authorrunning{J. Qi et al.} 
\maketitle

\begin{abstract}
Out-Of-Distribution generalization (OOD) is all about learning invariance against environmental changes. If the context\footnote[1]{In this paper, the word ``context'' denotes any class-agnostic attributes such as color, texture and background. The formal definition can be found in Appendix, A.2.} in every class is evenly distributed, OOD would be trivial because the context can be easily removed due to an underlying principle: \textbf{class is invariant to context}. However, collecting such a balanced dataset is impractical. Learning on imbalanced data makes the model bias to context and thus hurts OOD.
Therefore, the key to OOD is context balance. We argue that the widely adopted assumption in prior work---the context bias can be directly annotated or estimated from biased class prediction---renders the context incomplete or even incorrect. In contrast, we point out the ever-overlooked other side of the above principle: \textbf{context is also invariant to class}, which motivates us to 
consider the classes (which are already labeled) as
the varying environments\footnote[2]{The word ``environments''~\cite{irm} denotes the subsets of training data built by some criteria. In this paper, we take a class as an environment---our key idea.} to resolve context bias (without context labels). 
We implement this idea by minimizing the contrastive loss of intra-class sample similarity while assuring this similarity to be invariant across all classes. 
On benchmarks with various context biases and domain gaps, we show that a simple re-weighting based classifier equipped with our context estimation achieves state-of-the-art performance. We provide codes on Github\footnote[3]{https://github.com/simpleshinobu/IRMCon}.

\end{abstract}
\section{Introduction}
\label{sec:1}

\begin{figure}[t]
\centering
\includegraphics[width=4.5in]{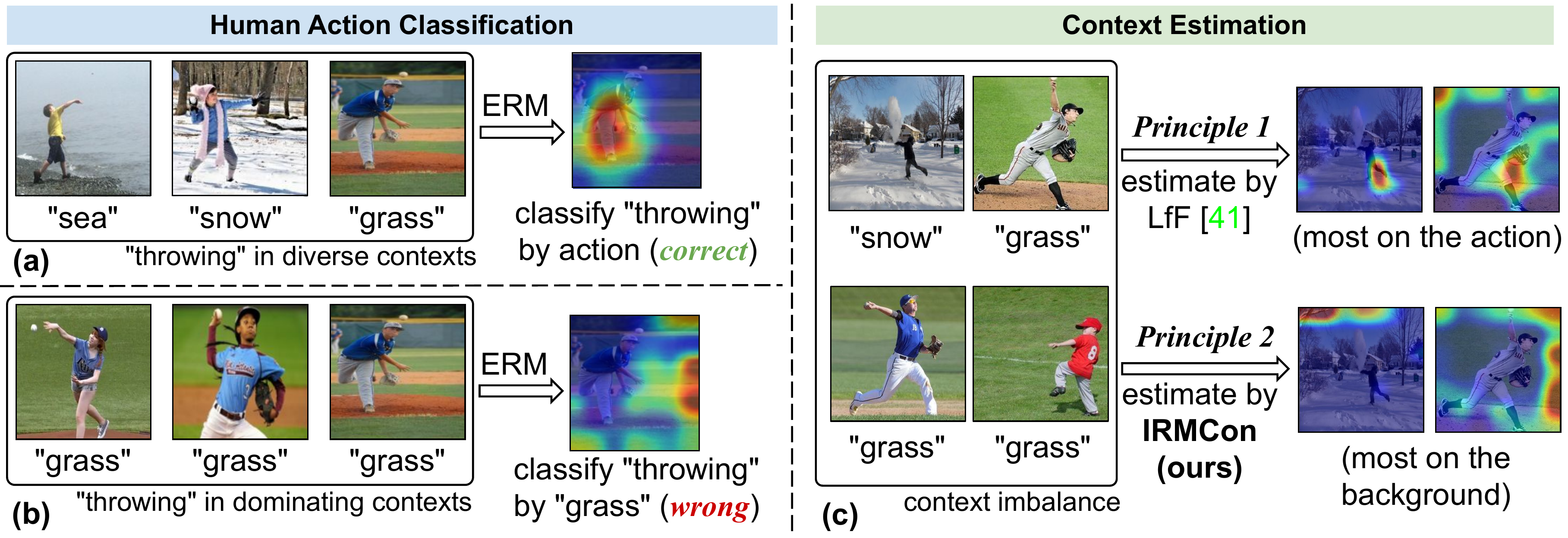}
\caption{GradCAM~\cite{selvaraju2017grad} visualizations of learned class and context. In (a) and (b): By using ERM, if the context is diverse and balanced within a class, the class feature is accurate---focused on the human's action; 
if the context dominates in the data, the class feature contains the context feature, e.g., the background ``grass''. 
In (c): The conventional context estimation~\cite{lff} based on Principle 1 is biased to class (focusing on the class of human action ``throwing''), while our IRMCon based on Principle 2 estimates better context (focusing on the background).}
\label{fig:teaser}
\end{figure}

The gold standard for collecting a supervised training dataset of quality is to ensure the samples per class are as diverse as possible and the diversities across classes are as evenly distributed as possible~\cite{deng2009imagenet,lin2014microsoft}. For example, the ``cat'' class should contain cats of varying contexts, such as types, poses, and backgrounds, and the rule also applies in the ``dog'' class. As illustrated in Fig.~\ref{fig:teaser}~(a),
on such a dataset, any 
Empirical Risk Minimization objective (ERM)~\cite{vapnik1992principles}, \eg, 
the widely used 
softmax cross-entropy loss~\cite{he2016deep}, 
can easily keep the class feature by penalizing inter-class similarities, while removing the context feature by favoring intra-class similarities. Thanks to the balanced context, the removal is clean. It can be summarized into the common principle:
\begin{manualtheorem}{1}
Class is invariant to context.
\end{manualtheorem}
\noindent For example, a ``cat'' sample is always a cat regardless of types, shapes, and backgrounds.

\begin{figure}[t]
\centering
\includegraphics[width=4in]{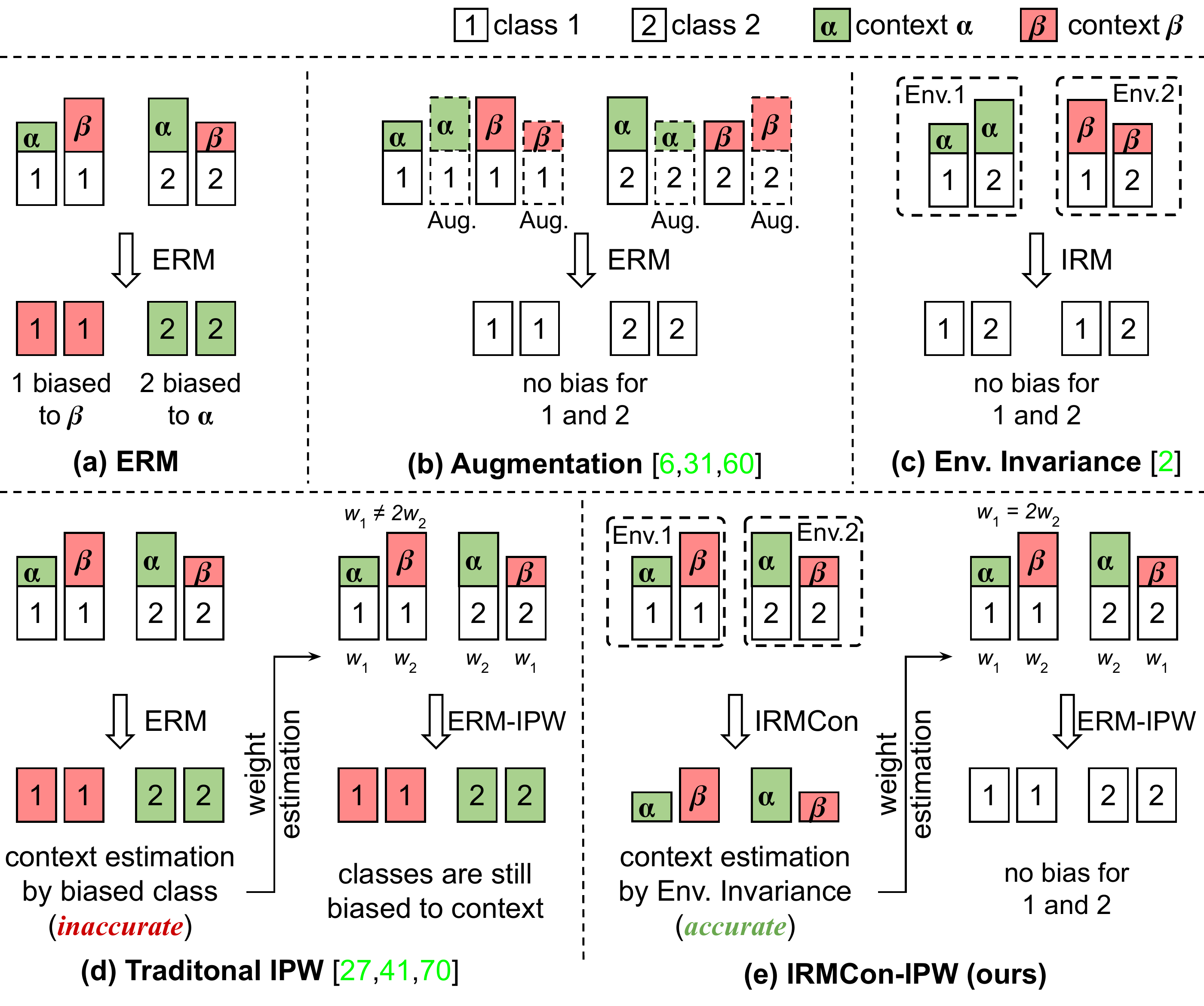}
\caption{Illustrations of the related approaches~\cite{irm,carlucci2019domain,feataug,li2018deep,lff,volpi2019addressing,zhang2021deep}. ERM is the baseline. Others and ours aim for mitigating context bias.
The components are elaborated below. 1) The length of a context bar indicates the number of samples in that context---longer bar means the context is more prevailing.
2) A sole bar with the mixture of a color and a class number denotes the feature biased to the prevailing context.
Our implementation method IRMCon-IPW is based on IRM and IPW, and our technical contribution (over the conventional methods of IRM or IPW) is the approach of disentangling context features not by using but by eliminating class features.
We provide a theoretical justification in Section~\ref{sec:4} and an empirical evaluation in Section~\ref{sec:5.2}.}
\label{fig:frameworks}
\end{figure}

Given testing samples whose contexts are Out-Of-(training)Distribution (OOD), the above ERM model can still classify correctly thanks to its focus only on the context-invariant class feature\footnote[1]{It is also known as causal or stable feature in literature~\cite{scholkopf2021toward,wang2021causal,zhang2021deep}.}---model generalization emerges~\cite{he2021towards,hendrycks2016baseline,liang2017enhancing}. However in practice, due to the limited annotation budget, real-world datasets are far from the ``golden'' balance, and learning the class invariance on imbalanced datasets is challenging. 
As shown in Fig.~\ref{fig:teaser}~(b), if the context ``grass'' in class ``throwing'' dominates the training, the model will use the spurious correlation ``most throwing actions happen in the grass'' to predict ``throwing''. 
Therefore, the obstacle to OOD generalization is context imbalance.

Existing methods for context or context bias estimation fall into two categories (details in Section~\ref{sec:2}). First, they annotate the context directly~\cite{irm,li2018deep}, as shown in Fig.~\ref{fig:frameworks} (c). 
This annotation takes additional costs.
Besides, it is elusive to annotate complex contexts.
For example, it is easy to label the coarse scenes ``water'' and ``grass'' but hard to further tell their fine-grained differences. Thus, context supervision is usually incomplete. 

Second, they estimate context bias by the biased class prediction~\cite{rebias,feataug,lff}, as shown in Fig.~\ref{fig:frameworks}~(d). 
This relies on the contra-position of Principle 1 which is essentially an \emph{indirect} context estimation.
\begin{manualtheorem}{1}
(Complement) If a feature is not invariant to context, it is not class but context.
\end{manualtheorem}
\noindent Here, the judgment of \emph{``not invariant to context''} is implemented by using the biased prediction of a classifier, \ie, if the classifier predicts wrongly, it is due to that the class invariance is not yet achieved in the classifier. 
Unfortunately, as the classifier is a combined effect of both class and context, it is ill-posed to disentangle if the bias is from biased context or immature class modeling. 
The reflection in the result is the incorrect context estimation mixed with class (see the upper part of Fig.~\ref{fig:teaser}~(c)). 
In fact, coinciding with recent findings~\cite{domainbed,wang2021causal}, we show in Section~\ref{sec:5} that existing methods with improper context estimation may even under-perform the ERM baseline. In particular, if the data is less biased, such methods may catastrophically mistake context for class---this limits their applicability only in severely biased training data.

In this paper, we propose a more \emph{direct} and accurate context estimation method without needing any context labels. Our inspiration comes from the other side of Principle 1:
\begin{manualtheorem}{2}
Context is also invariant to class.
\end{manualtheorem}
\noindent For example, the context ``grass'' is always grassy regardless of its foreground object class.

Principle 1 implies that the success of learning class invariance is due to the varying context. Similarly, Principle 2 tells us that we can learn context invariance with varying classes, and this is even easier for us to implement because the classes (taken as varying environments~\cite{irm}) have been labeled and balanced---a common practice for any supervised training data with an equal sample size per class. 
In Section~\ref{sec:4}, as illustrated in Fig.~\ref{fig:frameworks} (e), we propose a context estimator trained by minimizing the contrastive loss of intra-class sample similarity which is invariant to classes (based on Principle 2). In particular, the invariance is achieved by Invariant Risk Minimization (IRM)~\cite{irm} with our new loss term. We call our method \textbf{IRMCon} where \textbf{Con} stands for context.
Fig.~\ref{fig:teaser}~(c) illustrates that our IRMCon can capture better context feature.
Based on IRMCon, we can simply deploy 
a re-weighting method, \eg, \cite{little2019statistical}, to generate the balancing weights for different contexts---context balance is achieved.

We follow \textsc{DomainBed}~\cite{domainbed} for rigorous and reproducible evaluations, including 1) a strong Empirical Risk Minimization (ERM) baseline that is used to be mistakenly poor in OOD, and 2) a fair hyper-parameter tuning validation set. Experimental results in Section~\ref{sec:5} demonstrate that our IRMCon can effectively learn context variance and eventually improve the context bias estimation, leading to a state-of-the-art OOD performance. 
Our another contribution in experiments is we propose a non-pretraining setting for OOD. 
It is known that many conventional experiment settings with pretraining, especially using the ImageNet~\cite{deng2009imagenet}, have data leakage issues as mentioned in related works~\cite{wah2011caltech,xian2018zero}. We have an in-depth discussion on these issues in Section~\ref{sec:5.2}.

\section{Related Work}
\label{sec:2}
\noindent\textbf{OOD Tasks}. Traditional machine learning heavily relies on the Independent and Identically Distributed (IID) assumption for training and testing data. Under this assumption, model generalization emerges easily~\cite{vapnik1992principles}. However, this assumption is often violated by data distribution shift in practice---the Out-of-Distribution (OOD) problem causes the catastrophic performance drop~\cite{hendrycks2019benchmarking,recht2019imagenet}. In general, any test distribution unseen in training can be understood as OOD tasks, such as debiasing~\cite{clark2019don,geirhos2018imagenet,kim2019learning,li2019repair,wang2019learning}, long-tailed recognition~\cite{khan2017cost,liu2019large,tang2020long}, domain adaptation~\cite{ben2007analysis,gong2016domain,tzeng2017adversarial,yue2021transporting} and domain generalization~\cite{li2017deeper,shen2021towards,muandet2013domain}. In this work, we focus on the most challenging one, where the distribution shift is unlabelled (\eg, different from long-tailed recognition, where the shift of class distribution is known) and even unavailable (\eg, different from domain adaptation, where the OOD data is available). We leave other related tasks as future work.

\noindent\textbf{Invariant Feature Learning}. The invariant class feature can help the model achieve robust classification when context distribution changes. The prevalent methods are: 1) \textit{Data augmentation}~\cite{carlucci2019domain,li2018deep,volpi2019addressing,intermix}. They pre-define some augmentations for images to enlarge the available context distribution artificially. As the features are only invariant to the augmentation-related contexts, they cannot deal with other contexts out of the augmentation inventory. 2) \textit{Context Annotation}~\cite{irm,mmd,sun2016deep}. They split data by different context annotation into environments, and penalize the model by the feature shifts among different environments. As the features are only invariant to the annotated context, the inaccurate and incomplete annotations will impact their feature invariance. 3) \textit{Causal Learning}~\cite{mahajan2021domain,peters2016causal,pfister2019invariant,wang2021causal}. They learn the causal representations to capture the latent data generation process. Then, they can eliminate the context feature and pursue causal effect by intervention. These methods are essentially the re-weighting methods below in a causal perspective. 4) \textit{Reweighting}~\cite{feataug,lff,zhang2021deep}. They rebalance the context by re-weighting to help invariance feature learning. But, they improperly estimate the context weights by involving class learning into the context bias estimation. This inaccurate estimation problem severely influences the re-weighting and invariant feature learning. In contrast, IRMCon directly estimates the context without class prediction. The key difference is demonstrated in Fig.~\ref{fig:frameworks} (d) and (e): the output of our IRMCon does not contain class feature.

\section{Common Pipeline: Invariance as Class}
\label{sec:3}
Model generalization in supervised learning is based on the fundamental assumption~\cite{higgins2018towards,wang2021self}: any sample $x$ is generated from the two disentangled features (or independent causal mechanisms~\cite{suter2019robustly}), $x=g(\mathbf{x}_c, \mathbf{x}_t$), where $\mathbf{x}_c$ is the class feature, $\mathbf{x}_t$ is the context feature, $g(\cdot)$ is a generative function that transforms the two features in vector space to sample space (\eg, pixels). In particular, the disentanglement naturally encodes the two principles. To see this for Principle 1, if we only change the context of $x$ and obtain a new image $x'$, we have $\mathbf{x}_c = \mathbf{x}'_c$ but $\mathbf{x}_t\neq \mathbf{x}'_t$---class is invariant to context;  Principle 2 can be interpreted in a similar way. Therefore, we'd like to learn a feature extractor $\phi_c(x) = \mathbf{x}_c$ that helps the subsequent classifier to predict robustly across varying contexts.

\subsection{Empirical Risk Minimization (ERM)}
\label{sec:3.1}
If the training data per class is balanced and diverse, \ie, containing sufficient samples in different contexts, ERM has been theoretically justified that it can learn the class feature extractor $\phi_c(x)$ by minimizing a contrastive based loss such as softmax cross-entropy (CE) loss~\cite{wang2021self}:
\begin{equation}
\centering
\label{eq:1}
\mathcal{L}_{\operatorname{ERM}}(\phi_c, f) = \frac{1}{N}\sum\limits_{i=1}^N \operatorname{CE}(y_i, \hat{y}_i = f(\phi_c(x_i))),
\end{equation}
where $y_i$ is the ground-truth label of $x_i$ and $\hat{y}_i$ is the predicted label by the softmax classifier $f(\cdot)$. 

However, when the data is imbalanced and less diverse, ERM cannot learn $\phi_c(x) = \mathbf{x}_c$. We illustrate this in Fig.~\ref{fig:frameworks} (a): if more class \texttt{1} samples contain context $\beta$ than $\alpha$, the resultant $\phi_c(x)$ will be biased to the prevailing context, \eg, features for classifying class \texttt{1} will be entangled with context $\beta$. To this end, augmentation-based methods~\cite{carlucci2019domain,volpi2018generalizing} aim to compensate for the imbalance (Fig.~\ref{fig:frameworks} (b)). However, as contexts are complex, augmentation will be far from enough to compensate for all of them.

\subsection{Invariant Risk Minimization (IRM)}
\label{sec:3.2}
If context annotation is available, we can use IRM~\cite{irm} to learn $\phi_c$ by applying Principle 1 that $\phi_c$ should be invariant to different contexts. Compared to ERM on balanced data that achieves invariance in a passive way via random trials~\cite{austin2011introduction}, 
IRM on imbalanced data adopts the active intervention, taking contexts as the environments:
\begin{equation}
\centering
\begin{split}
\mathcal{L}_{\operatorname{IRM}}(\phi_c, \theta) = \sum_e \frac{1}{|e|}\sum\limits_{(x_i,y_i)\in e} \left[\operatorname{CE}(y_i,\hat{y}_i) + \lambda \|\nabla_{\theta} \operatorname{CE} (y_i,\hat{y}_i^{\theta})\|^2\right],
\end{split}
\label{eq:2}
\end{equation}
where $\hat{y}_i^{\theta}=f(\phi_c(x_i)\cdot \theta)$, $e$ is one of the environments of the training data according to context labels, and $\lambda>0$ is a trade-off hyper-parameter for the invariance regularization term. 
$\theta$ is a dummy classifier, whose gradient is not applied to update itself but to calculate the regularization term in Eq.~\eqref{eq:2}.
The regularization term encourages $\phi_c$ to be equally optimal in different environments, \ie, become invariant to environments (contexts). We follow IRM~\cite{irm} to set $\theta$ as $1$.

As illustrated in Fig.~\ref{fig:frameworks} (c), if we want to learn a common classifier that discriminates \texttt{1} and \texttt{2} in both environments, the only way is to remove the context $\alpha$ and $\beta$. However, it has been demonstrated by~\cite{liu2021heterogeneous,wang2021causal} that the context annotation
is usually incomplete and using it may even under-perform ERM.

\subsection{Inverse Probability Weighting (IPW)}
When context annotation is unavailable, we can estimate the context and then re-balance data according to context. We begin with the following ERM-IPW loss~\cite{causal-erm,seaman2018introduction}: 
\begin{equation}
\centering
\begin{split}
\mathcal{L}_{\operatorname{ERM-IPW}}(\phi_c,\phi_t, f) =  \frac{1}{N}\sum\limits_{i=1}^N \operatorname{CE}(y_i, \hat{y}_i = f(\phi_c(x_i)))\cdot \frac{1}{P(x_i\mid \phi_t(x_i))}.
\end{split}
\label{eq:3}
\end{equation}
We can see that the key difference between ERM-IPW and ERM is the sample-level IPW term $1/P(x_i|\phi_t(x_i))$, where $\phi_t(x) = \mathbf{x}_t$ is the context feature extractor. This IPW implies that if $x$ is more likely associated with its context $\mathbf{x}_t$, \ie, the class feature counterpart $\mathbf{x}_c$ is also more likely associated with $\mathbf{x}_t$, we should under-weight the loss because we need to discourage such a context bias.

\begin{wrapfigure}{R}{60mm}
\centering
\includegraphics[width=2.1in]{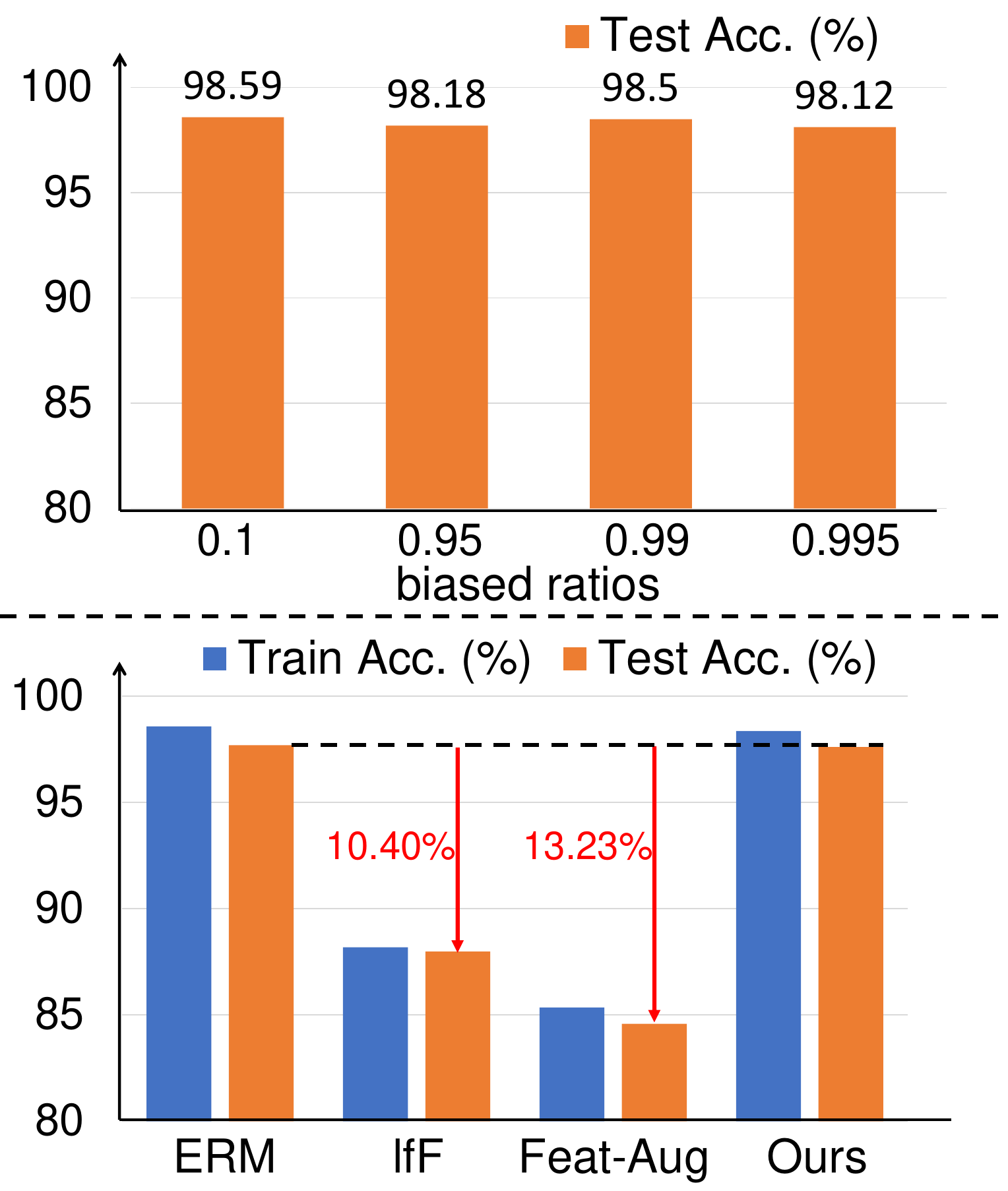}
\caption{Accuracy (\%) of models when training on \textit{Colored MNIST} context-balance set. \textbf{Top}: ERM is stable in test sets with varying context biases; \textbf{Bottom}: 
due to the incorrect context estimation, traditional reweighting methods degenerate significantly compared to ERM when training on context-balance set. 
Thanks to the correct context estimation, our IRMCon-IPW achieves comparable performance to ERM.
}
\label{fig:bar}
\end{wrapfigure}

However, the context estimation of $\phi_t$ is almost challenging as learning $\phi_c$. Instead, a prevailing strategy is to estimate it by a biased classifier~\cite{feataug,lff}, \eg, 
\begin{equation}
\centering
\begin{split}
P(x|\phi_t(x))\propto
\frac{\operatorname{CE}(y,\hat{y} = f(\phi_c(x))) + \operatorname{CE}(y,\hat{y} = f_b(\phi_b(x)))}{\operatorname{CE}(y,\hat{y} = f_b(\phi_b(x)))},
\end{split}
\label{eq:4}
\end{equation}
where $\phi_b$ is the bias feature extractor and $f_b$ is the bias classifier. $\phi_b$ and $f_b$ are minimized by ERM equipped with generalized cross entropy (GCE) loss~\cite{zhang2018generalized}:
\begin{equation}
\centering
\begin{split}
\mathcal{L}_{\operatorname{ERM}}(\phi_b, f_b) = \frac{1}{N}\sum\limits_{i=1}^N \operatorname{GCE}(y_i, \hat{y}_i = f_b(\phi_b(x_i))),
\end{split}
\label{eq:5}
\end{equation}
where $\operatorname{GCE}(y,\hat{y})\!=\!\sum_{k=1}^n y_k\cdot \frac{1-{\hat{y}_k}^q}{q}$ is used to amplify the bias, where $q$ is a constant, $k$ is the index of class and $n$ is the class number. However, the loss in Eq.~\eqref{eq:5} inevitably includes the effect from the class feature $\mathbf{x}_c$, due to the aforementioned assumption $x = g(\mathbf{x}_c, \mathbf{x}_t)$. In other words, such a combined effect cannot distinguish whether the bias is from class or context, resulting in inaccurate context estimation. 
We show the illustration in Fig.~\ref{fig:frameworks} (d).
Specifically, the weights are estimated from class and context, and thus inaccurate to balance the context.
In addition, the experimental results in Fig.~\ref{fig:bar} (Bottom) testify that: inaccurate context estimation will severely hurt the performance, \ie, fail to derive unbiased classifiers.

\section{Our Approach: Invariance as Context}
\label{sec:4}

To tackle the inaccurate context estimation of $\phi_t(x)$, we propose to apply Principle 2 as a way out. As illustrated in Fig.~\ref{fig:frameworks} (e), if we consider each class as the environment, we can clearly see that the \emph{unique} environmental change is the class which has been already labeled. This motivates us to apply IRM to learn invariance as context by removing the environment-equivariant class. The crux is how to design the contrastive based loss---more specifically, how to modify $\theta$ and $\operatorname{CE}(\cdot)$ in Eq.~\eqref{eq:2}. The following is our novel solution.

We design a new contrastive loss based on the intra-class (environment) sample similarity, as follows,
\begin{equation}
\centering
\begin{split}
\mathcal{L}_{ct}(\phi_t,e,\theta) = \sum_{x_i \in e} -log \frac{exp(\phi_t(x_i)^T \phi_t(\texttt{Aug}{(x_i)})\cdot \theta)}{\sum_{x_i' \in e} exp(\phi_t(x_i)^T \phi_t(x_i')\cdot \theta)},
\end{split}
\label{eq:6}
\end{equation}
where $\texttt{Aug}(\cdot)$ is the common augmentations, such as flip and Gaussian noise (used in standard contrastive 
losses~\cite{chen2020simple,grill2020bootstrap,he2020momentum}), $e$ is the environment split by class, \eg, under the environment $e_1$, any $x_i \in e_1$ has the class label \texttt{1}, $\theta$ is the dummy classifier, we add $\theta$ here for the convenience to introduce Eq.~\eqref{eq:7}.
The reason for using contrastive loss is that it preserves all the intrinsic features of each sample~\cite{oord2018representation,wang2021self}. Yet, without the invariance to class, $\phi_t(x)\neq \mathbf{x}_t$. Then, based on Eq.~\eqref{eq:2}, our proposed IRMCon for learning ``invariance as context'' is:
\begin{equation}
\centering
\begin{split}
\mathcal{L}_{\operatorname{IRMCon}}(\phi_t,\theta) =
\sum_e \frac{1}{|e|}[\mathcal{L}_{ct}(\phi_t,e,\theta) + \lambda  |\nabla_{\theta}\mathcal{L}_{ct}(\phi_t,e,\theta)|],
\end{split}
\label{eq:7}
\end{equation}
where $\theta$ plays the same role in Eq.~\eqref{eq:2}, to regularize $\phi_t$ be invariant to environments (classes).
We can prove that solving Eq.~\eqref{eq:7} achieves $\phi_t(x)= \mathbf{x}_t$,
\ie, the context feature is disentangled (see Appendix).
As demonstrated in Fig.~\ref{fig:context_visualization}, $\phi_t$ can extract accurate context features. 
Thanks to $\phi_t$, we can further improve IPW:
\begin{equation}
\centering
\begin{split}
P(x|\phi_t(x))\propto
\frac{\operatorname{CE}(y,\hat{y} = f(\phi_c(x))) + \operatorname{CE}(y,\hat{y} = f_b(\mathbf{x}_t))}{\operatorname{CE}(y,\hat{y} = f_b(\mathbf{x}_t))},
\end{split}
\label{eq:8}
\end{equation}
where $\mathbf{x}_t=\phi_t(x)$. We train $f_b$ by using GCE loss, just replacing $\phi_b(x)$ with $\mathbf{x}_t$ in Eq.~\eqref{eq:5}. $\phi_t$ is trained by IRMCon and then fixed when estimating the context. 
\begin{figure}[t]
\centering
\includegraphics[width=4.5in]{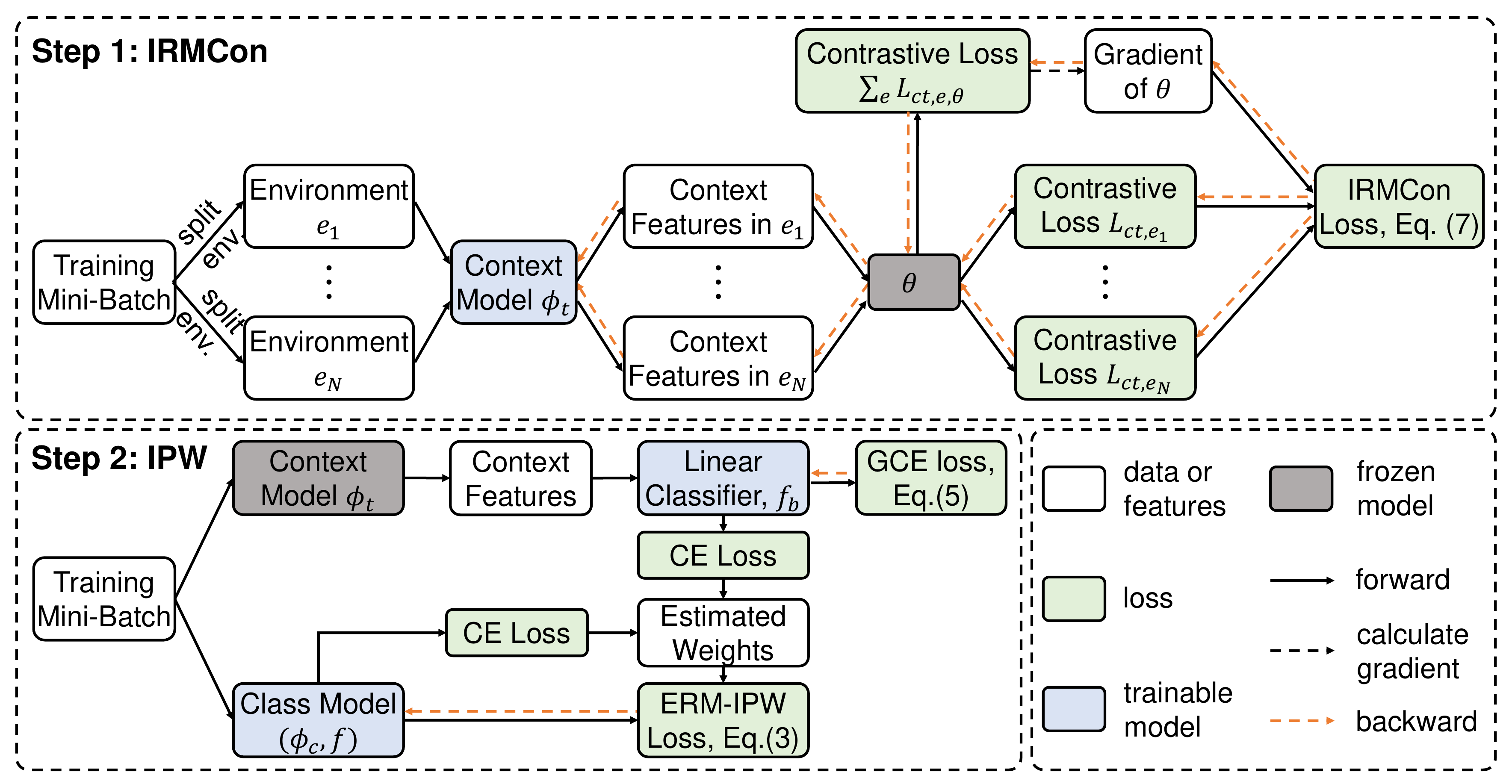}
\caption{
The training pipeline of our IRMCon-IPW. 
1) ``split env.'' denotes we split the training samples in mini-batch into subsets based on class labels, i.e., samples of each class in one subset, forming $N$ environments $\{e_i\}_1^N$; %
2) $\theta$ is a dummy classifier, whose gradient is for regularizing $\phi_t$ become invariant to classes. See the detailed algorithm in Appendix}
\label{fig:pipeline}
\end{figure}

As shown in
Fig.~\ref{fig:reweighting_comparasion}, our biased classifier can estimate more accurate weights to perform better reweighting than the traditional one
We streamline the proposed IRMCon-IPW in Fig.~\ref{fig:pipeline} and summarize our algorithm in Appendix.

\begin{figure}[t]
\centering
\includegraphics[width=4.3in]{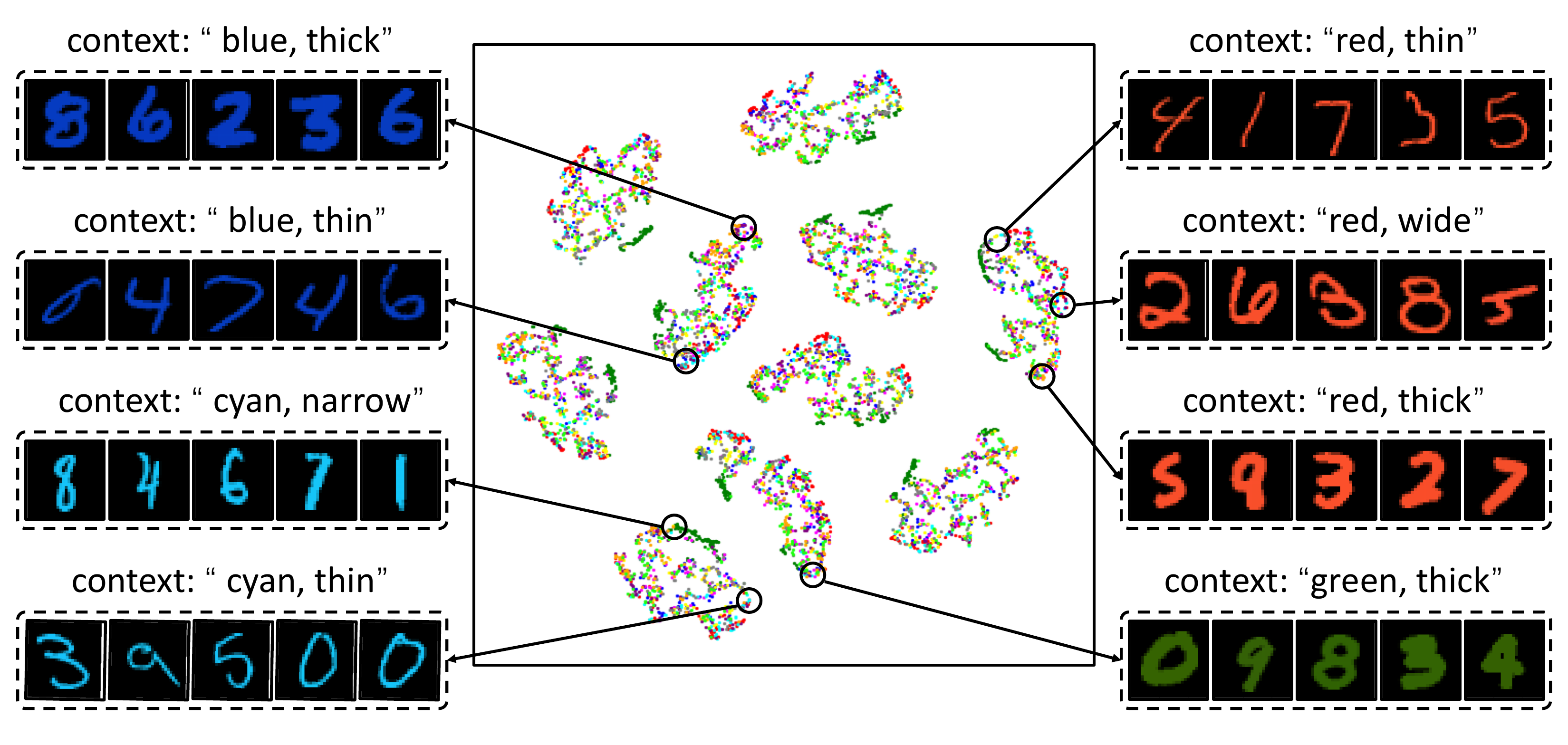}
\caption{t-SNE~\cite{van2008visualizing} visualizations of our context features of the \textit{Colored MNIST} test samples. The color of points denotes their class labels. IRMCon is trained on the 99\% biased training set.  Features are naturally clustered by context. As there is no context ground-truth, the context labels are interpreted by us.}
\label{fig:context_visualization}
\end{figure}

\section{Experiments}
\label{sec:5}

We introduce the benchmarks of two OOD generalization tasks, removing context bias (also called debias) and mitigating domain gaps (also called domain generalization and termed DG), and our implementation details in Section~\ref{sec:5.1}. 
Then, we evaluate the effectiveness of our approach based on the experimental results in Section~\ref{sec:5.2}.

\begin{table}[t]
\centering
\caption{Accuracy (\%) on context biased datasets compared with SOTA methods. We reproduced the methods and averaged the results over three independent trials (mean$\pm$std). ``*'': For reproducing mismatch issues, performance is quoted from the original paper. Our reproduced results are reported in Appendix. ``-'': no report in that setting.}

\scalebox{0.95}

{
\begin{tabular}{l   c  c  c  c  c  c  c }
\Xhline{2\arrayrulewidth}
\multirow{3}{*}{Dataset}&\multirow{3}{*}{\makecell[c]{Bias\\ Ratio(\%)}}&
\multicolumn{6}{c}{Methods} \\
\cline{3-8}
& & \multirow{2}{*}{ERM} & \multirow{2}{*}{\makecell[c]{Rebias\\~\cite{rebias}}} & \multirow{2}{*}{\makecell[c]{EnD$^*$\\~\cite{tartaglione2021end}}} & \multirow{2}{*}{\makecell[c]{LfF\\~\cite{lff}}} & \multirow{2}{*}{\makecell[c]{Feat-Aug$^*$\\~\cite{feataug}}} & \multirow{2}{*}{\makecell[c]{IRMCon-IPW \\(Ours)}}\\
\\
\hline
\multirow{6}{*}{\rotatebox[origin=c]{90}{\makecell[c]{Colored\\ MNIST}}}& 99.9 &
20.4\scriptsize$\pm$1.1&20.8\scriptsize$\pm$0.6&- &{56.8}\scriptsize$\pm$1.6&-&\textbf{66.7}\scriptsize$\pm$2.3\\
&99.8 &
26.4\scriptsize$\pm$0.4&28.3\scriptsize$\pm$0.9&-&{68.3}\scriptsize$\pm$1.5&-&\textbf{75.5}\scriptsize$\pm$1.5\\
&99.5 &
42.9\scriptsize$\pm$1.1&44.4\scriptsize$\pm$0.5 &34.3\scriptsize$\pm$1.2&{77.0}\scriptsize$\pm$1.5&65.2\scriptsize$\pm$4.4&\textbf{81.0}\scriptsize$\pm$0.9\\
&99.0 &
59.2\scriptsize$\pm$0.5&58.6\scriptsize$\pm$0.4  &49.5\scriptsize$\pm$2.5&{82.5}\scriptsize$\pm$1.7&81.7\scriptsize$\pm$2.3&\textbf{85.3}\scriptsize$\pm$0.3\\
&98.0 &
72.5\scriptsize$\pm$0.2&73.5\scriptsize$\pm$1.0  &68.5\scriptsize$\pm$2.2&84.1\scriptsize$\pm$1.5&{84.8}\scriptsize$\pm$1.0&\textbf{88.3}\scriptsize$\pm$0.2\\
&95.0 &
85.7\scriptsize$\pm$0.5&85.5\scriptsize$\pm$0.5 &81.2\scriptsize$\pm$1.4 &86.8\scriptsize$\pm$0.5&{89.7}\scriptsize$\pm$1.1&\textbf{92.2}\scriptsize$\pm$0.5\\
\hline
\multirow{4}{*}{\rotatebox[origin=c]{90}{\makecell[c]{Corrupted\\ Cifar-10}}}
&99.5& 22.7\scriptsize$\pm$0.5
& 22.7\scriptsize$\pm$0.7
&22.9\scriptsize$\pm$0.3
& 26.1\scriptsize$\pm$0.7
& 30.0\scriptsize$\pm$0.7
& \textbf{31.0}\scriptsize$\pm$0.6\\
&99.0 & 25.8\scriptsize$\pm$0.6
& 24.9\scriptsize$\pm$0.7
& 25.5\scriptsize$\pm$0.4
& 31.8\scriptsize$\pm$0.7
& 36.5\scriptsize$\pm$1.8 
& \textbf{37.1}\scriptsize$\pm$0.4\\
&98.0 & 28.7\scriptsize$\pm$0.1
& 29.1\scriptsize$\pm$0.7
& 31.3\scriptsize$\pm$0.4
& 38.9\scriptsize$\pm$1.0 
& 41.8\scriptsize$\pm$2.3  
& \textbf{42.5}\scriptsize$\pm$1.0\\
&95.0 & 39.9\scriptsize$\pm$1.6
& 38.9\scriptsize$\pm$1.7
& 40.3\scriptsize$\pm$0.9
& 51.3\scriptsize$\pm$0.9
& 51.1\scriptsize$\pm$1.3 
& \textbf{53.8}\scriptsize$\pm$1.3\\
\hline
\multirow{2}{*}{\rotatebox{90}{\makecell[c]{\\BAR}}}
&99.0 & {52.9}\scriptsize$\pm$0.7 &  52.1\scriptsize$\pm$0.5&-& 48.1\scriptsize$\pm$2.7 &52.3\scriptsize$\pm$1.0 & \textbf{55.3}\scriptsize$\pm$0.6\\
&95.0 & {65.2}\scriptsize$\pm$1.9&  65.0\scriptsize$\pm$1.8&-&60.6\scriptsize$\pm$2.6& 63.5\scriptsize$\pm$1.5&\textbf{67.9}\scriptsize$\pm$0.8\\
\Xhline{2\arrayrulewidth}
\end{tabular}
}

\label{table:1}
\end{table}

\subsection{Datasets and Settings}
\label{sec:5.1}

\noindent\textbf{Context Biased Datasets}. 
We follow LfF~\cite{lff} to use two synthetic datasets,

\begin{wrapfigure}{R}{65mm}
\centering
\includegraphics[width=2.2in]{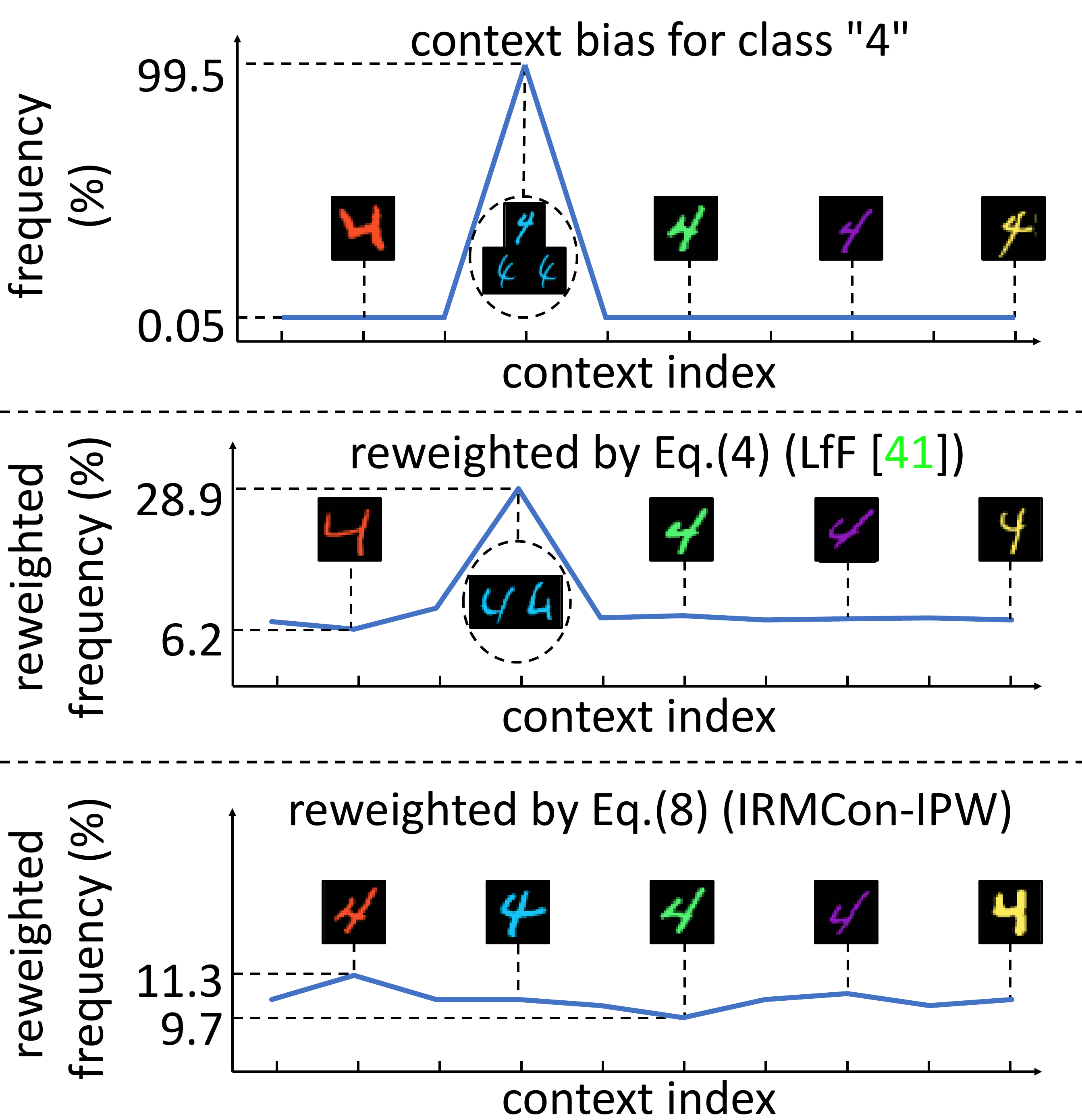}
\caption{Illustrations of the reweighted sample frequencies for 10 color contexts. 
All models are trained on the 99.5\% biased \textit{Colored MNIST}.
The reweighted frequency of a context indicates the normalized sum over the inverse probabilities of the samples in this context.
\textbf{Top}: 
Biased context distribution in the training set. \textbf{Middle}: Biased context distribution derived by using LfF~\cite{lff}. 
\textbf{Bottom}: Relatively balanced context distribution by using our method.
}
\label{fig:reweighting_comparasion}
\end{wrapfigure}
\noindent\textit{Colored MNIST} and \textit{Corrupted CIFAR-10}, and one real-world dataset, \textit{Biased Action Recognition} (\textit{BAR})~\cite{lff} for evaluation.

On each dataset, we manually control the context bias ratio by generating (in synthetic datasets) or sampling (in the real-world dataset) training images. 

In specific, on \textit{Colored MNIST}, we follow LfF to generate 10 colors as 10 contexts.
We connect each digit (class) with a specific color and dye them with the ratio from
\{99.9\%, 99.8\%, 99.5\%, 99.0\%, 98.0\%, 95.0\%\} to construct each biased training set.
In the test set, 10 colors are uniformly distributed on the samples of each class. 
For \textit{Corrupted CIFAR-10}, we follow LfF to use \{Saturate, Elastic, Impulse, Brightness, Contrast, Gaussian, Defocus Blur, Pixelate, Gaussian Blur, Frost\} as 10 contexts. 
Similar to \textit{Colored MNIST}, we generate context biased training set by pairing a context and a class with a ratio chosen from \{99.5\%, 99.0\%, 98.0\%, 95.0\%\}.
In the test set, 10 corruptions are uniformly distributed.

The real-world dataset \textit{BAR}
contains six kinds of action-place bias, and each one is between human action and background, e.g., ``throwing'' always happens with the ``grass'' background;
We choose a bias ratio in \{99.0\%, 95.0\%\}. 

\noindent\textbf{Domain Gap Dataset}.
We use \textit{PACS}~\cite{li2017deeper} to testify our method. It consists of seven object categories spanning four image domains: \emph{Photo}, \emph{Art-painting}, \emph{Cartoon}, and \emph{Sketch}.
We follow~\textsc{DomainBed}~\cite{domainbed} to each time select three domains for training and the left one for testing.
More details about datasets, \eg, the number and size of the training images, are given in Appendix. 

\noindent\textbf{Comparing Methods}. As the two types of datasets have their own state-of-the-art (SOTA) methods, we compare with different SOTA methods in context biased benchmark and domain gap benchmark, respectively.

For context biased datasets, we compare with Rebias~\cite{rebias}, End~\cite{tartaglione2021end}, LfF~\cite{lff}, and Feat-Aug~\cite{feataug}. 
For domain gap dataset (DG task), we compare with domain-label based methods, such as DANN~\cite{dann}, fish~\cite{fish}, and TRM~\cite{TRM}, as well as domain-label free methods, such as RSC~\cite{RSC} and StableNet~\cite{zhang2021deep}. 
As we claimed at the end of Section~\ref{sec:3.1}, we train all models from scratch.
This makes some DG methods (\eg, MMD~\cite{mmd} and CDANN~\cite{cdann}) hard to converge.

\begin{table}[t]
\centering
\caption{Accuracy (\%) on the domain generalization dataset \textit{PACS}~\cite{li2017deeper}. We reproduced all the methods by the \textsc{DomainBed}~\cite{domainbed} code base without pretraining. Results are averaged over 3 independent trials (mean$\pm$std). ``-'' denotes that methods fail to converge when training from scratch.}
\scalebox{0.95}
{
\begin{tabular}{l l c c  c  c  c}
\hline
\multicolumn{2}{c}{\multirow{2}{*}{Methods}}&
\multicolumn{5}{c}{PACS}\\
\cline{3-7}
\multicolumn{2}{l}{}  &\emph{Art.} & \emph{Cartoon}  & \emph{Photo} & \emph{Sketch} & Avg.  \\
\hline

\multirow{8}{*}{\rotatebox[origin=c]{90}{\makecell[c]{w/ domain \\ supervision}}}&IRM\cite{irm}& 31.1\scriptsize$\pm$1.4 & 38.7\scriptsize$\pm$2.5 & - & 44.4\scriptsize$\pm$2.2  &  - \\

&DRO~\cite{dro}& 39.0\scriptsize$\pm$1.9 & 53.8\scriptsize$\pm$1.2 & 63.6\scriptsize$\pm$2.9 & \textbf{62.4}\scriptsize$\pm$0.6 & 54.7\\

&InterMix~\cite{intermix}& {42.2}\scriptsize$\pm$0.5 & 52.8\scriptsize$\pm$1.9 &61.0\scriptsize$\pm$2.4& 58.4\scriptsize$\pm$1.0 & 53.6\\

&MLDG~\cite{mldg}&38.8\scriptsize$\pm$0.7 & 53.5\scriptsize$\pm$0.7 &63.3\scriptsize$\pm$0.1& 60.2\scriptsize$\pm$1.2&  54.0\\

&DANN~\cite{dann} &31.5\scriptsize$\pm$1.1 & 48.2\scriptsize$\pm$1.6 &58.1\scriptsize$\pm$1.5& 44.9\scriptsize$\pm$0.7& 45.7\\

&V-REx~\cite{vrex} & 33.9\scriptsize$\pm$1.2 & 40.9\scriptsize$\pm$1.2 & - & 55.1\scriptsize$\pm$2.9 &-\\

&Fish~\cite{fish} & \textbf{43.1}\scriptsize$\pm$2.1 & \textbf{57.4}\scriptsize$\pm$0.4 & \textbf{64.8}\scriptsize$\pm$2.7 & 61.1\scriptsize$\pm$0.8& \textbf{56.6}\\

&TRM~\cite{TRM} & 41.8\scriptsize$\pm$1.8& 54.9\scriptsize$\pm$0.8 & - & 61.3\scriptsize$\pm$2.3 & - \\

\hline
\multirow{6}{*}{\rotatebox[origin=c]{90}{\makecell[c]{w/o  domain \\ supervision}}}&ERM& 40.4\scriptsize$\pm$0.7 & 54.3\scriptsize$\pm$0.3 & 63.7\scriptsize$\pm$0.4 & 58.9\scriptsize$\pm$2.6 & 54.3\\

&SD~\cite{sd} & 39.1\scriptsize$\pm$0.8 & 54.4\scriptsize$\pm$1.4 & 61.7\scriptsize$\pm$3.8&51.3\scriptsize$\pm$3.2 & 51.6\\

&RSC~\cite{RSC}&{40.7}\scriptsize$\pm$1.1&49.8\scriptsize$\pm$6.0&58.0\scriptsize$\pm$1.9&53.3\scriptsize$\pm$4.3&50.5\\

&LfF~\cite{lff}&38.2\scriptsize$\pm$1.4 & 50.4\scriptsize$\pm$0.9 &58.0\scriptsize$\pm$0.6& 60.4\scriptsize$\pm$1.2& 51.8\\

&IRMCon-IPW& \textbf{40.9}\scriptsize$\pm$1.7 & \textbf{56.0}\scriptsize$\pm$2.9 & \textbf{64.9}\scriptsize$\pm$0.7 & \textbf{61.1}\scriptsize$\pm$2.5 & \textbf{55.7}\\
\hline
\end{tabular}
}

\label{table:2}

\end{table}

\noindent\textbf{Implementation Details}. We first introduce two implementation details to deal with the implementation issues we met, and then provide training details. 

\noindent1) \emph{Weighted sample strategy}. This strategy is for the biased dataset. For example, under the 99.9\% biased training set, in a mini-batch, all the images may have the same context in a class, unless we can sample over 1,000 images per class to get 1 sample with non-biased context. To solve this issue, we use the bias model from LfF~\cite{lff} to learn an inaccurate context estimator, and based on its inverse probability we sample a relative context-balanced mini-batch.
This strategy frees us from sampling a very large batch to learn Eq.~\eqref{eq:6}.

\noindent2) \emph{Strategy for learning augmentation-related context}. 
It is hard to learn augmentation related context, when using contrastive loss. To minimize contrastive loss, the model needs to learn invariance on augmentations, \ie, augmentation related features will be removed.
On \textit{Corrupted Cifar-10}, we add the classification loss in Eq.~\eqref{eq:5} to our IRMCon loss to train the context extractor. Please note that we use this strategy only for \textit{Corrupted Cifar-10} as context on this dataset is dominated by augmentation-related context, such as 95\% ``car'' has augmentation-related context `Gaussian noise''. Due to space limits, we put other details in Appendix.

\noindent3) \emph{Training details}. On the \textit{Colored MNIST}, we use 3-layers MLPs to model $\phi_c,\phi_b$ and $\phi_t$. On the \textit{Corrupted Cifar-10}, we use ResNet-18 for $\phi_c$ and 3-layers CNNs for $\phi_b$ and $\phi_t$.
On the \textit{BAR} and \textit{PACS}, we use ResNet-18 for $\phi_c,\phi_b$ and $\phi_t$. For optimization in context biased datasets, we follow LfF~\cite{lff} to use Adam~\cite{kingma2014adam} optimizer with the learning rate as 0.001. Other detailed settings, \eg batch size, epochs, and $\lambda$ in each setting, can be found in Appendix.


On all datasets, we follow \textsc{DomainBed}~\cite{domainbed} to randomly split the original unbiased test set into 20\% and 80\% as the validation set and test set, respectively, and select the best model based on validation results. 
We average the results of three independent runs, and report them in the format of ``mean accuracy $\pm$ standard deviation''.

\subsection{Results and Analyses}
\label{sec:5.2}
\noindent\textbf{IRMCon-IPW achieves SOTA.} We show our results of context biased datasets in Table~\ref{table:1} and domain gap dataset in Table~\ref{table:2}.

1) Table~\ref{table:1} presents that our IRMCon-IPW achieves very clear margins over the related methods.
In particular, the improvements are more obvious in the settings of higher bias ratios. The possible reason is when the bias ratio is higher, the ``rare'' context samples become less. 
Reweighting methods are more sensitive to the accuracy of context weights estimation. Therefore, accurate context estimation plays a more essential role.
Compared to related methods, our IRMCon can estimate more accurate context, \ie, extract high-quality context features like the illustration in Fig.~\ref{fig:context_visualization}, whose gain over others is more obvious when increasing the context bias ratio.

2) Table~\ref{table:2} presents that on the domain gap dataset, our method outperforms ERM and also achieves the best average performance over all the domain label-free methods. In addition, it achieves comparable results to 
the other DG methods (in the upper block) which need domain labels.

\begin{wrapfigure}{R}{55mm}
\centering
\includegraphics[width=2.1in]{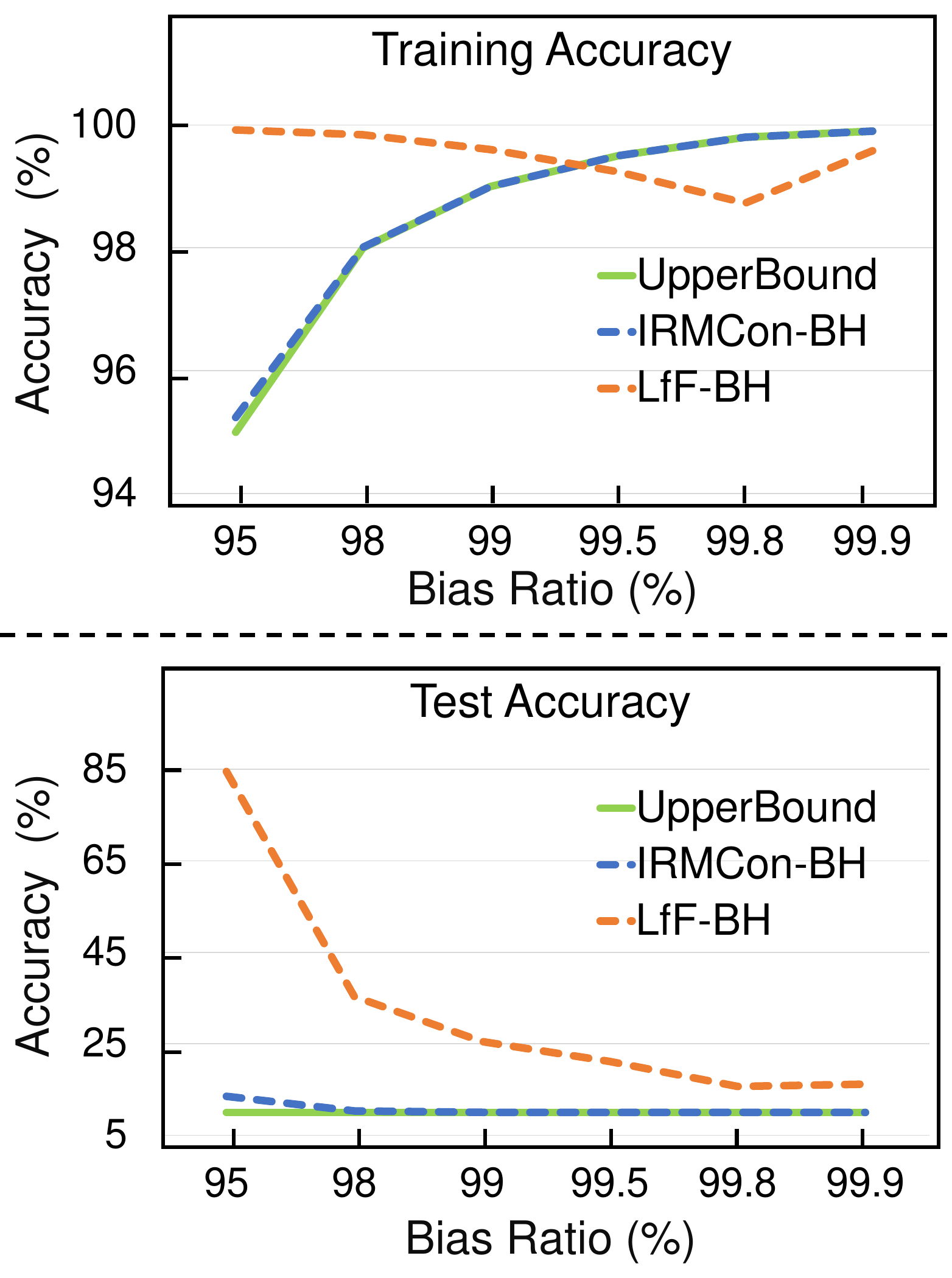}
\caption{
Comparing the bias classification heads in LfF~\cite{lff} (LfF-BH) and in ours (IRMCon-BH) on \textit{Colored MNIST} with different bias ratios. The bias classification heads (BH) intentionally use context to predict class. Our bias head is almost the same as the upper bound case in test set---random class prediction (10\%).}
\label{fig:biasmodel_line_chart}
\end{wrapfigure}
\noindent\textbf{Why does ERM perform so well in most cases?} %
On \textit{PACS}, we follow the \textsc{DomainBed}~\cite{domainbed} to implement a strong ERM baseline.
On \textit{BAR}, we use the strong augmentation strategy, Random Augmentation~\cite{cubuk2020randaugment}, which can be considered as an OOD method as shown in Fig.~\ref{fig:frameworks} (b).
If we do not apply such strong augmentations, ERM performance drops significantly. We show the corresponding results in Appendix.

\noindent\textbf{Why do we train models from scratch for OOD problems?} 
We challenge the traditional pretraining settings in some OOD tasks, such as Domain Generalization, because we are concerned that the data or knowledge of the test set has been leaked to the model when pretrained on large-scale image datasets. Data leakage is a usual problem in pretraining settings, such as ImageNet~\cite{deng2009imagenet} leaks to CUB~\cite{wah2011caltech}. Such problem will severely destroy the validity of the OOD task~\cite{xian2018zero}. Empirically, we provide an observation in Domain Generalization to justify our challenge. In pretraining settings, ERM achieves the ``impressive'' 98\% test accuracy~\cite{domainbed}
when \emph{Photo}
domain is used for testing. This number is significantly higher (around 20\% higher) than using \emph{Cartoon} and \emph{Sketch} in testing. However, this is not the case if there is no pretraining on ImageNet, see Table~\ref{table:2}, bottom block first line, ERM method.
The reason is that ImageNet, collected from the real world, leaks more real images in \emph{Photo}, compare to artificial images in \emph{Cartoon} and \emph{Sketch}.
Therefore, we propose the non-pretraining setting for all OOD benchmarks to prevent the leakage problem.

\noindent\textbf{How to evaluate the context feature learned in IRMCon-IPW?}
We visualize the comparisons between the context features learned by IRMCon-IPW and LfF in Fig.~\ref{fig:biasmodel_line_chart}.
We show the training and test accuracies of the linear classifiers (we call bias classification heads) that are trained with context features and class labels, \ie, to learn the bias intentionally.
We can see from the figures that ours shows the almost same learning behavior as the upper bound case: context is invariant to class and should predict class by random chance.
It means that IRMCon-IPW is able to recover the oracle distribution of contexts in the image. 
This can be taken as a support to the bottom illustration in Fig.~\ref{fig:reweighting_comparasion} where using our weights can achieve a balanced context distribution---the ground truth distribution.

\begin{figure}[t]
\centering
\includegraphics[width=4.4in]{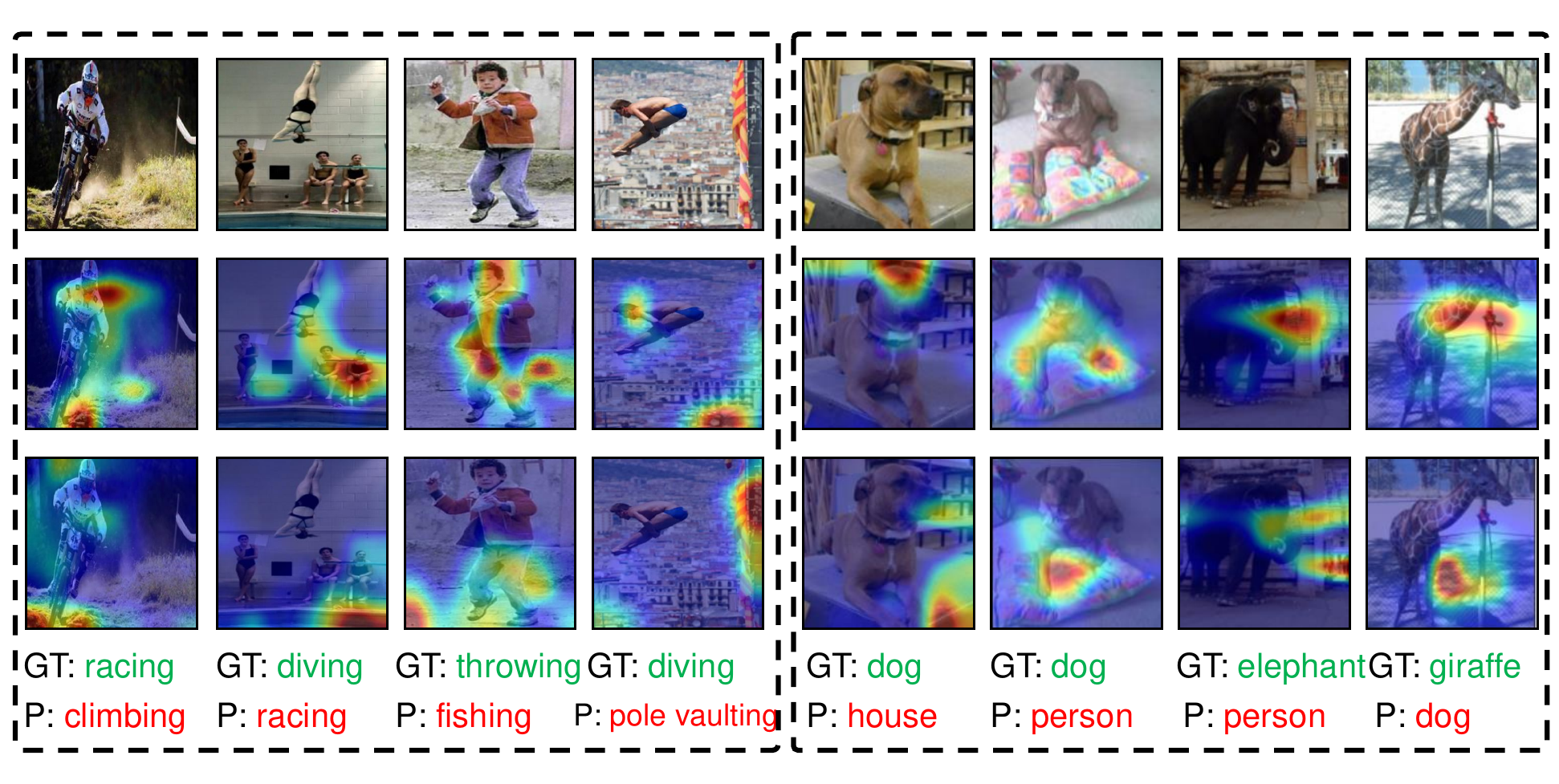}
\vspace{-5pt}
\caption{GradCAM~\cite{selvaraju2017grad} visualizations of IRMCon-IPW failure cases. \textbf{Top}: input test images; \textbf{Middle}: context visualization by bias classifier of IRMCon; \textbf{Bottom}: class visualization. Left four columns are selected from \textit{BAR} test set, the model is trained on the 99\% biased training set; right four are selected from the \emph{Photo} domain of \textit{PACS}, model is trained on the other three domains. GT: ground-truth label; P: predicted label.}
\label{fig:failure_samples}
\end{figure}

\noindent\textbf{How does IRMCon-IPW tackle domain gap issues?}
Compared to the datasets with pre-defined context distribution in training (\eg, set color distribution in each class in \textit{Colored MNIST} dataset~\cite{lff}), the domain gap dataset such as \textit{PACS} does not have such explicit context settings. While it has implicit context distribution related to the domain. 
This distribution is often imbalanced which leads to context bias problems (similar to context biased datasets such as \textit{BAR}).
Therefore, our method can help \textit{PACS} to ``debias''. 
We notice that, compared to ERM, our improvement for \textit{PACS} is not as significant as that on the context biased datasets.  This might be because the context bias in \textit{PACS} is not as severe as that in context biased datasets. 

\noindent\textbf{Failure cases.} We show some failure cases of our IRMCon in Fig.~\ref{fig:failure_samples}. The failure cases are selected if their IRMCon-IPW classification results are wrong. As expected, we see that the key reasons for failure are the incorrect context estimation, \eg, the contexts are mixed with the foreground or wrongly attended to the foreground. By inspecting the \textit{BAR} dataset, we find that some contexts, \eg, ``pool'' for the class ``diving'', are relatively unique for certain classes. This implies that the context is NOT invariant to class. To resolve this, we conjecture that this is a dataset failure and the only way out is to bring external knowledge.

\section{Conclusions}
Context imbalance is the main challenge in learning class invariance for OOD generalization. Prior work tackles this challenge in two ways: 1) relying on context supervision and 2) estimating context bias by classifier failures. We showed how they fail and hence proposed a novel approach called IRM for Context (IRMCon) that directly learns the context feature without context supervision.  The success of IRMCon is based on: \emph{context is invariant to class}, which is the overlooked other side of the common principle---class is invariant to context. Thanks to the class supervision which has been already provided as environments in training data, IRMCon can achieve context invariance by using IRM on the intra-class sample similarity contrastive loss. We used the context feature for Inverse Probability Weighting (IPW): a method for context balancing, to learn the final classifier that generalizes to OOD. IRMCon-IPW achieves state-of-the-art results on several OOD benchmarks.

\noindent\textbf{Acknowledgements}
This research is supported by the National Research Foundation, Singapore under its AI Singapore Programme (AISG Award No: AISG2-RP-2021-022), Alibaba-NTU Singapore Joint Research Institute (JRI) and Alibaba Innovative Research (AIR) programme, A*STAR under its AME YIRG Grant (Project No.A20E6c0101).

\bibliographystyle{splncs04}
\bibliography{egbib}


\end{document}